# Deep Learning For Computer Vision Tasks: A review

Rajat Kumar Sinha
Department of Computer Science and Engineering
IIIT Bhubaneswar
Odisha, India, 751003
Email: rajatsinha333@gmail.com

Ruchi Pandey
Department of Communication Engineering and Signal Processing
IIIT Bhubaneswar
Odisha, India, 751003
Email: ruchipandey295@gmail.com

Rohan Pattnaik
Department of Computer Science and Engineering
IIIT Bhubaneswar
Odisha, India, 751003
Email: b114031@iiit-bh.ac.in

*Abstract*—Deep learning has recently become one of the most popular sub-fields of machine learning owing to its distributed data representation with multiple levels of abstraction. A diverse range of deep learning algorithms are being employed to solve conventional artificial intelligence problems. This paper gives an overview of some of the most widely used deep learning algorithms applied in the field of computer vision. It first inspects the various approaches of deep learning algorithms, followed by a description of their applications in image classification, object identification, image extraction and semantic segmentation in the presence of noise. The paper concludes with the discussion of the future scope and challenges for construction and training of deep neural networks.

## I. INTRODUCTION

Amongst numerous sub-fields of machine learning, deep learning is the one where learning happens with high-level data using hierarchical structures. It improves chip programming abilities on low cost computing hardware by applying advanced machine learning algorithms. In recent years several attempts have been made towards the advancement of deep learning algorithms. In most of the observations it has been realized that deep learning based approach provides more satisfactory results as compared to numerous other state-of-the-art schemes. In spite of a series of remarkable accomplishments, deep learning still remains a really young field. Keeping it in mind, this paper surveys the recent advances in deep learning and the application of these algorithms in the field of computer vision. The rest of the paper is organized as follows: Section 2, provides an overview of the different techniques of deep learning like Convolutional Neural Networks(CNN), Recurrent neural networks(RNN), Restricted Boltzmann Machines(RBM), Autoencoders and Extreme Learning. Section 3, presents the potential applications of these techniques along with some comparisons. The challenges of deep learning approaches encountered during constructing and training the data are discussed in Section 4. Finally, Section 5 gives the concluding remarks.

## II. TECHNIQUES

Deep learning is a key area of research in the field of Image and Video processing, Computer vision [2,3] and Bio-informatics to name a few. This gave rise to the introduction and application of several variants of deep learning in the above mentioned fields. However, the techniques of deep learning generally are divided into three categories namely Convolutional Neural Networks(CNN), Restricted Boltzmann Machines(RBM) and Autoencoders. Additionally Recurrent Neural Networks and Extreme Learning are also a few techniques frequently used in this field. For better clarity, the architectural descriptions along with layer information and detailed functionalities of these techniques are described in a nutshell.

### A. Convolutional Neural Network (CNNs)

When it comes to substantial training of multiple layers, the Convolutional Neural Network(CNN) is considered as the most momentous approach for a variety of applications [4]. The CNN architecture shown in the fig. 1, consists of mainly three types of layers, namely [1]-convolutional layers, pooling layer and fully connected layers. Training the network can be divided into forward and backward stages. In the forward stage first we classify the input image depending upon its weights and bias for each layer. The loss cost is calculated from the input data by using the predicted output. In the backward stage, depending on the loss cost measured, the gradients are calculated for each parameter. Using the gradients, it then updates the parameters for the next iteration. The training procedure can be halted after satisfactory number of repetitions. The functionalities of the said network is described as follows:

*1) Convolutional Layers:* In this layer, the CNN uses numerous filters to convolve the entire image including intermediate feature maps & generating different feature maps.

The major privileges [5] of the convolution operation are-

- Reduce number of parameters using weight sharing mechanisms.
- Correlation between neighboring pixels are easy due to local connectivity.
- Location of object is fixed.

This advantages leads researchers to replace fully connected layers to put forward the learning process [6,14].

*2) Pooling Layers:* This layer is similar to convolution layer but minimizes the measurements of feature maps and also the parameters of the network. Generally average and max pooling are used. For average and max pooling, Boureu et al. [11] have demonstrated the theoretical details about their performances. Among all the three layers, the pooling layer happens to be the most profusely investigated. There exist mainly three approaches, with varied usage, that are related to the pooling layers. All of these different pooling approaches are discussed as follows:

- **Stochastic pooling**- To overcome the problem of sensitivity for over-fitting the training set in max pooling [13], stochastic pooling is introduced. Here the activation in each pooling area is allocated randomly as per a multinomial distribution. In this method, similar kind of inputs are considered with very small variation to overcome the problem of over-fitting.
- **Spatial pyramid pooling**- As the input dimensions of an image is generally fixed while using CNN, the accuracy is compromised in case of of a variable sized image. This problem can be eliminated[15] by replacing the last layer of CNN architecture with spatial pyramid pooling layer. It takes an arbitrary input and gives a flexible solution with respect to size, aspect ratio, scales.
- **Def pooling**- Another type of pooling layer, that is sensitive to the distortion of visual patterns, known as def pooling is used for solving the ambiguity of deformation in computer vision.

*3) Fully-connected layers:* The final layer of CNN consists of 90% of the parameters. The feed forward network forms a vector of a particular length to follow up processing. Since these layers contain most of the parameters, there is a high computational burden while training the data.

### B. Recurrent neural networks (RNNs)

Unlike CNN the Recurrent neural networks (RNNs) are generally used to deal with sequential data to represent time dependencies for various applications such as translating natural languages, music, time-series data, handwriting recognition, video processing etc. Deep convolutional RNN is a good candidate for frame prediction and object classification. As sequential data is trained using RNN, this network involves large number of complex parameters to train as compared to other neural networks. The architecture, optimization and training is complex in RNN model in order to train the sequential steps.

As Hidden Markov Model depends upon the previous data it seems impractical to model the long time dependencies data. This problem is eliminated by RNN as it deals with the present data only with high accuracy. Now compared to other neural networks RNN gives good results for image captioning and analysis.

### C. Restricted Boltzmann Machines(RBMs)

Minton et al. have in [11], proposed a generative random neural network technique called the Restricted Boltzmann machine. In RBM (a form of the Boltzmann Machine), the hidden and visible layer units are required to form a bipartite graph. As the RBM architecture is a double barreled graph, the hidden layer units H & the visible layer units $V_1$ display conditional independence. Therefore,

$$P(HV_1)=P(H_1V_1)P(H_2V_1)...P(H_nV_1)$$

In the above equation, given $V_1$ as input, 'H' can be derived using $P(HV_1)$. Likewise, $V_1$ can be obtained through $P(HV_1)$. By altering the parameters, we can reduce the difference between $V_1$ & $V_2$, and resulting 'H' will give us a fine lineament of $V_1$.

The practical implication along with detailed analysis is proposed by Minton [12]. It consists of flexible learning of an intensified gradient to resolve those problems. Because information passes via numerous layers of feature extractor, the model estimates the binary units corrupted by noise to conserve learned features. Many deep learning models [16] can be constructed by applying RBMs as learning models like Deep Boltzmmann Machine(DBMs), Deep Belief Networks(DBNs) and Deep Energy Models(DEMs).

Their description & their application to the field of Computer vision are respectively given as:

*1) Deep Boltzmann Machines(DBMs):* Like the RBM, the DBMs are also the part of the Boltzmann family. It consists of several layers of masked elements, where the odd and even-numbered layers are conditionally independent of each other. While training DBMs, all the supervised models are jointly trained. This type of training yields significant improvements in case of both likelihood [20] and classification.

A major drawback of DBM is that the computational time requirement for approximate inference is comparatively greater than a Deep Belief Network, which in case of big datasets, renders the combined optimization of the parameters of Deep Boltzmann Machines, non-viable.

*2) Deep Belief Networks(DBNs):* Hinton et al. [17], introduced the deep belief network. This model is a

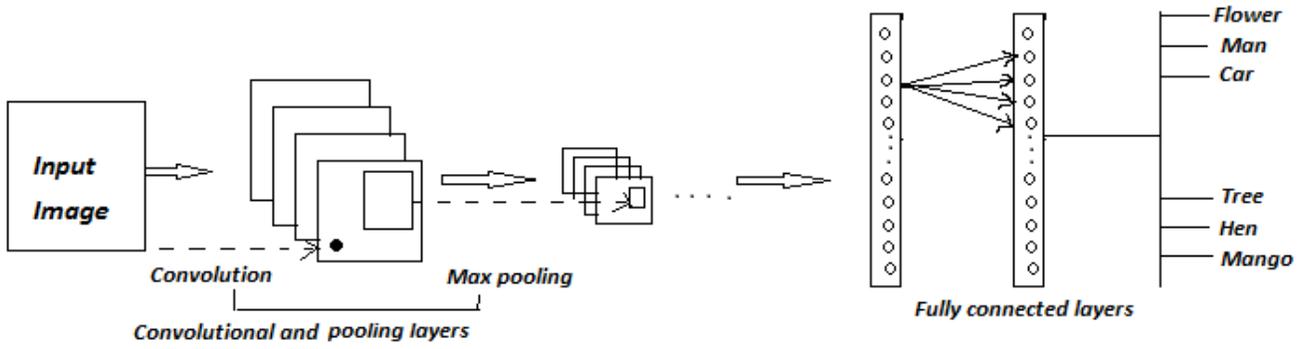

Fig. 1: Pipelined Architecture of Convolutional Neural Network

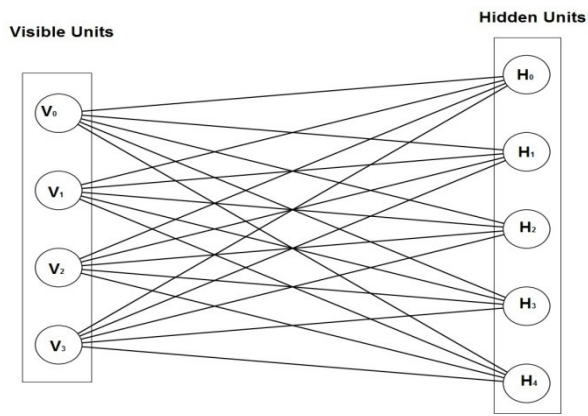

Fig. 2: Architecture of Restricted Boltzmann Machines

probabilistic productive model that facilitates a collective expectation distribution over distinguishable labels and data. It is a layer-by-layer training approach which has the following benefits:

- Parameter selection is done for generation of appropriate initialization of the network, leading to a poor local optima till a certain level.
- Since it is a type of unsupervised learning, it makes a conclusion depending upon the clustered data without the need for any labelled data required for training. Although, it is a computationally exorbitant approach.

*3) Deep Energy Models(DEMs):* Ngiam et al. [19], introduced a deep Energy model, which is considered to be an approach to train the deep architectures. DEM is superior than DBMs and DBNs as it eliminates the issue of sharing the feature of having manifold random masked layers by only having one layer of random masked units. Thus DEM facilitates systematic training and a good conclusion about the data.

Although, CNNs are widely used over RBMs for many vision related tasks. Keeping this in mind, in order to shape both the local and global structure in face segmentation, Kae et al.[18], merged the CRF and the RBM, severely decreasing the error in face labelling.

*D. Autoencoders*

A notable variant of artificial neural networks, the autoencoder is used for training logical encodings [21]. The autoencoder learns to refurnish its own inputs instead of predicting any target value 'T' for given input 'S' by training the network accordingly. The learned feature is obtained by optimizing the reconstruction error and the correspondingcode which uses back propagation variant with same dimensionality. Some variants of autoencoders are presented as follows:

- **Sparse autoencoder-** It is used to [22,23] extract scattered features from input data. Sparse auto encoder is used. Sparse auto encoders are used as (i)For high dimension representation it is easy to divide it in different classes depending upon their likelihood same as SVM. (ii)We can interpret the complex input data as we have many different classes depending upon their likelihood.(iii)Sparse encoders can be used for biological vision.
- **De-noising autoencoder (DAE)-** It is used to [24,25] amplify the resilience of the given model as it is able to work in the presence of random noise.
- **Contractive autoencoder (CAE)-** It is [26] the advanced version of DAE in which durability is achieved by adding contractive penalty to optimize reconstruction error used for unsupervised and transfer learning challenges.

*E. Extreme Learning*

A very recent topic in the vast areas of machine learning happens to be Extreme learning given by Guang-Bin Huang. It is a feed-forward neural network used for regression and classification tasks. It consists of a solitary layer of masked nodes in which the weights which are assigned as inputs to the masked nodes are random and are never corrected. In one step, the weights between the masked nodes and outputs are learned, which leads to learning of a linear model. These are better than networks trained by using

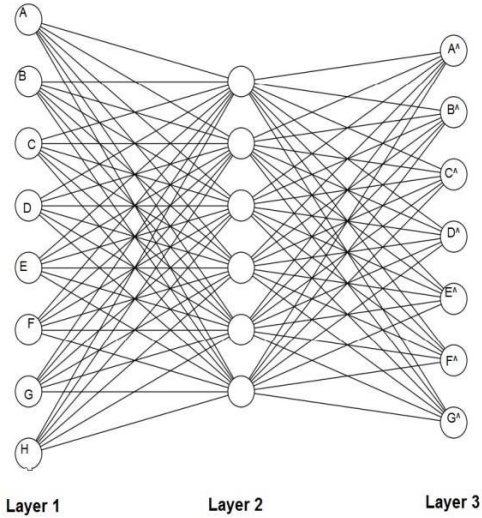

Fig. 3: Architecture of Autoencoder

back-propagation because of their faster learning ability and a good generalization capability.

**Algorithm:** Extreme learning Machine training algorithm learns a model of the following kind:

$\hat{Y} = W_2 \sigma(W_1 x)$

Where $W_1$ is the matrix comprising of weights of the input-to-hidden layer, $\sigma$ is an activation function and $W_2$ is matrix of hidden to output layer weights. Given a design matrix 'X':

- $W_1$ should be filled with Gaussian Random Noise.
- $W_2$ should be estimated by least-square-fit to a matrix of response variable 'Y', with the pseudo inverse computed.

$W_2 = \sigma(W_1 X)^+ Y$

## III. APPLICATIONS

Deep learning is a recent variant of artificial neural network. There are several potential applications of deep learning. Here we have presented some of the most notable applications related to image and video processing.

### A. Image Classification

Depending upon the probability of its presence visual object class labeling can be done [27]. In deep learning most of the methods involve bags of visual words through which initially we get a histogram of quantized visual words then we proceed for classification. Most of the time sparse coding is used to recover the information loss.

### B. Object Identification

Object identification is somewhat related to image classification. In object identification process, it takes image as an input and then object estimation can be done depending upon the class and positions. The PASCAL VOC with 20 classes is often used for object identification. Deep learning (CNN based) is an advanced method after Deformable Part model (DPM) [28] for object identification. It is applicable for all scales unlike DPM. By using deep learning, we can divide the images into different classes using classifiers(SVM) and then check the presence of an object. Recurrent Convolutional Neural Networks(RCNNs) are used for fast object identification.

### C. Image Retrieval

Image retrieval involves the collection of images having same object. Many CNN models are used and give good performance compared to traditional methods like VLAD and Fisher vector. Derived from the idea of Spatial Pyramid to extract patches for multiple scales Reverse SVM [29] is used. Without eliminating any spatial information it takes the complete image and then divides it into different scales. Using CNN models a large dataset can be trained and used for feature extraction and can also be applied for image retrieval to achieve better accuracy.

### D. Semantic Segmentation

CNN models are used for semantic segmentation tasks, as it is potent of handling the pixel-level predictions. Output masks having a 2-dimensional spatial spread are required by semantic segmentation. The process of semantic segmentation is outlined below:

- **Detection based Segmentation:** This approach divides the images depending on object detection [30,31,32,33]. Recurrent Convolutional Neural Network along with simultaneous detection and segmentation generated the proposal for object identification and then applying traditional approach to divide the regions and to assign the pixels as activation vectors, which resulted in large improvements. Disadvantage of this approach is additional expense for object identification. Convolutional feature masking is an approach in which we proposals are personally obtained using the feature maps.
- **FCN-CRFs Based Segmentation:** In this approach, the fully connected layer is replaced by the fully convolutional layers. It is a commonly used technique for semantic segmentation [34,35]. A similar model is proposed by FCN integrated with conditional random fields(CRFs) into FCN [36] for elaborate boundary retrieval.
- **Weakly supervised annotations:** Other than improving the segmentation model, Weakly supervised segmentation is another area of interest for researchers [37]. Here it takes the segmentation of training data depending upon which it estimates the segmentation masks to update the neural network. Then it combines all small pixel levels so that we can get the complete image.

## IV. CONCLUSION

In this paper we have given an survey of Deep learning and its recent development. The analysis of prevailing deep learning architectures is done by developing a categorical layout. Deep learning algorithms are divided into three categories: Convolutional Neural Network, Restricted Boltzmann Machines, Autoencoder. Apart from that, RNN and extreme learning are also quite popular. In this paper we mainly dealt with the recent advancement of CNN dependent strategies, since it is mostly used for images. In some recent papers it has been reported CNN dependent strategies have surpassed the precision of human testers. In spite of the many researches reported till now, there is a large scope for additional advancements. As an example in theoretical foundation, it has not yet been shown that when and under what conditions they will perform well and an optimal structure for certain tasks. This paper summarizes and gives an idea to the new researchers to explore more in the vast yet young area of Deep learning.